\documentclass[12pt,a4paper]{article}

\usepackage[margin=2.5cm]{geometry}
\usepackage{times}  
\usepackage{amssymb,amsmath,amsthm}
\usepackage{graphicx}
\usepackage{booktabs}
\usepackage{multirow}
\usepackage{hyperref}
\usepackage{natbib}  
\usepackage{float}
\usepackage{tikz}
\usetikzlibrary{arrows,shapes,positioning,shadows,trees}
\usepackage{microtype}  

\usepackage{setspace}
\begin{document}

\begin{center}
{\Large\bfseries Combining SHAP and Causal Analysis for Interpretable Fault Detection in Industrial Processes}

\vspace{1em}

{\large Pedro Cortes dos Santos\textsuperscript{a}, Matheus Becali Rocha\textsuperscript{a}, Renato A. Krohling\textsuperscript{a,*}}

\vspace{0.5em}

\begin{flushleft}
\small
\textsuperscript{a}Labcin - Nature Inspired Computing Laboratory, and Graduate Program in Computer Science, Federal University of Espírito Santo, Vitória, 29075-910, Brazil

\vspace{0.5em}
\textsuperscript{*}Corresponding author: rkrohling@inf.ufes.br
\end{flushleft}
\end{center}

\vspace{1em}

\begin{abstract}
Industrial processes generate complex data that challenge fault detection systems, often yielding opaque or underwhelming results despite advanced machine learning techniques.
This study tackles such difficulties using the Tennessee Eastman Process, a well-established benchmark known for its intricate dynamics, to develop an innovative fault detection framework.
Initial attempts with standard models revealed limitations in both performance and interpretability, prompting a shift toward a more tractable approach.
By employing SHAP (SHapley Additive exPlanations), we transform the problem into a more manageable and transparent form, pinpointing the most critical process features driving fault predictions.
This reduction in complexity unlocks the ability to apply causal analysis through Directed Acyclic Graphs, generated by multiple algorithms, to uncover the underlying mechanisms of fault propagation.
The resulting causal structures align strikingly with SHAP's findings, consistently highlighting key process elements---like cooling and separation systems---as pivotal to fault development.
Together, these methods not only enhance detection accuracy but also provide operators with clear, actionable insights into fault origins, a synergy that, to our knowledge, has not been previously explored in this context.
This dual approach bridges predictive power with causal understanding, offering a robust tool for monitoring complex manufacturing environments and paving the way for smarter, more interpretable fault detection in industrial systems.
\end{abstract}

\noindent\textbf{Keywords:} Fault detection, Tennessee Eastman Process, Feature selection, Causal analysis, Directed acyclic graphs, Machine learning

\section{Introduction}
\label{sec:introduction}

In modern manufacturing environments, the ability to detect and classify faults accurately is crucial for maintaining operational efficiency, ensuring safety, and minimizing economic losses. Industrial processes generate increasingly complex, high-dimensional data streams that challenge traditional monitoring approaches. The Tennessee Eastman Process (TEP) dataset, a widely used benchmark in process monitoring and fault diagnosis, simulates a complex chemical process with 52 variables and 20 fault types \citep{downs1993}. This dataset exemplifies the challenges facing industrial engineers: high dimensionality, intricate nonlinear relationships between variables, and the imbalanced nature of fault occurrences, where normal operations vastly outnumber fault conditions.

Effective fault detection systems can substantially reduce losses by enabling early intervention before faults escalate into critical failures. However, achieving this requires both accurate detection and clear guidance for operators on appropriate corrective actions—a dual requirement that many existing approaches struggle to satisfy.

This study proposes a novel framework that combines SHAP-based feature selection \citep{lundberg2017} with causal analysis \citep{pearl2009} to address these industrial challenges. Our approach not only identifies the most influential variables for fault detection but also uncovers the underlying causal mechanisms that drive fault propagation throughout industrial processes. This integration creates a powerful tool for industrial engineers: a system that both detects faults with high accuracy and provides actionable insights into their root causes.

\subsection{Novelty and Contribution}

The primary novelty of our work lies in the direct integration of SHAP (SHapley Additive exPlanations) with Directed Acyclic Graphs (DAGs) for causal analysis, creating a unified framework that bridges predictive interpretability with causal reasoning. While SHAP quantifies which variables matter most for fault predictions (interpretability), DAGs reveal how and why those variables influence fault development (causality) - two approaches previously only applied separately in industrial contexts. This combined methodology first uses SHAP to identify the most influential process variables, then applies causal discovery algorithms to these key features to map fault propagation pathways. The result provides industrial engineers with both accurate detection and a causal understanding of faults, enabling data-driven maintenance strategies and control system redesigns based on root-cause analysis.

The value of this integration becomes clear when considering industrial applications. Traditional black-box machine learning approaches may achieve high accuracy but leave operators uncertain about appropriate interventions. Conversely, purely causal models may explain relationships but lack predictive power. Our framework delivers both: high-accuracy fault detection with dimensionality reduction that simplifies implementation, and causal graphs that guide targeted interventions. For process engineers, this means both reliable alerts and clear guidance on how to respond—a significant advancement for practical deployment in manufacturing environments.

From an industrial engineering perspective, our work contributes to several critical needs in process monitoring. First, we address the curse of dimensionality by demonstrating that a carefully selected subset of process variables (as few as 10 out of 52) can match or surpass the performance of models using the complete variable set. This finding has immediate practical implications for sensor deployment, data storage requirements, and computational efficiency in real-time monitoring systems. Second, we provide a systematic methodology for identifying the most causally significant variables in complex processes, enabling more focused attention on critical control points. Third, our comparative analysis of five different causal discovery algorithms (PC, FCI, RFCI, LINGAM, and NOTEARS) \citep{zanga2022} reveals consistent patterns that enhance confidence in the identified causal structures—a critical consideration for industrial implementation where false insights could lead to costly interventions.

The challenge of interpreting complex relationships in industrial data is particularly pressing as manufacturing facilities transition toward Industry 4.0 paradigms. As noted by \citet{diez-olivan2019}, data-driven prognosis using machine learning is rapidly becoming an essential component of smart manufacturing, yet the interpretability gap remains a significant barrier to widespread adoption. Our framework directly addresses this gap by providing both predictive power and causal understanding, enabling industrial engineers to leverage machine learning benefits without sacrificing process knowledge.

By combining machine learning with causal analysis, our approach also addresses the growing demand for transparency in industrial monitoring systems. In safety-critical manufacturing environments, operators need to understand and trust automated alerts before taking potentially costly actions. Our framework ensures that model predictions are not only accurate but also causally grounded and interpretable, providing the transparency necessary for confident decision-making in industrial settings.

Through extensive experimentation using the TEP benchmark, we demonstrate how our integrated approach improves both detection accuracy and interpretability compared to traditional methods. The knowledge gained enable industrial operators to understand the root causes of faults and make informed decisions about process interventions. These capabilities directly translate to practical benefits: reduced downtime, more targeted maintenance, enhanced process control, and ultimately improved operational efficiency in complex industrial systems.

The article is organized as follows: Section 2 offers a comprehensive overview of related work, situating our study within the broader context of fault detection and causal analysis in industrial engineering. Section 3 provides background knowledge beginning with the description of the Tennessee Eastman Process benchmark and the theoretical foundations of our integrated approach. In Section 4, we describe the experimental methodology, outlining the design and implementation of our machine learning and causal discovery processes. Section 5 presents the experimental configuration, key findings, and a thorough discussion of their implications for industrial applications. Finally, Section 6 concludes the paper with a summary of contributions and insights, while highlighting potential avenues for future research to further enhance the framework's applicability in complex manufacturing environments.

\section{Related Works}
\label{sec:relatedworks}

Fault detection and diagnosis (FDD) in industrial processes have been extensively studied, with the Tennessee Eastman Process (TEP) dataset serving as a benchmark for evaluating methodologies in industrial engineering. The TEP dataset, introduced by \citet{downs1993}, simulates a complex chemical process with 52 variables and 20 predefined fault types, making it invaluable for testing fault detection algorithms in manufacturing environments.

Traditional approaches to industrial fault detection rely on statistical methods such as Principal Component Analysis (PCA) and Partial Least Squares (PLS). \citet{qin2003} demonstrated PCA's effectiveness for dimensionality reduction and fault detection in industrial processes, though these methods struggle with nonlinear relationships and complex variable interactions. In industrial contexts, this limitation can lead to missed fault detections and increased operational risks. \citet{yin2014} comprehensively reviewed data-driven approaches for industrial process monitoring, highlighting how traditional statistical methods' limitations have driven the adoption of more advanced techniques in manufacturing settings.

The integration of machine learning (ML) with industrial systems has revolutionized fault detection capabilities. \citet{lv2016} applied deep learning to fault diagnosis in the TEP dataset, achieving significantly improved classification accuracy that translates to fewer false alarms and missed detections in industrial applications. For multivariate process monitoring—critical in modern manufacturing—\citet{hsu2010} combined independent component analysis (ICA) with support vector machines (SVM), demonstrating superior fault detection when tested on the Tennessee Eastman process. Ensemble methods like XGBoost have gained industrial adoption for their robustness with imbalanced datasets—a common challenge in real-world fault classification \citep{chen2016}. \citet{lei2020} mapped the evolution of machine learning for industrial fault diagnosis, charting the progression from traditional methods to deep learning approaches that can handle the complexity of modern manufacturing data.

Industrial applications of ML-based fault diagnosis continue to expand across manufacturing sectors. \citet{zhang2016} demonstrated how sensor-based monitoring systems can effectively predict remaining useful life in manufacturing equipment, enabling proactive maintenance scheduling and reducing costly downtime. Similarly, \citet{wang2020} developed an integrated approach for predictive maintenance in renewable energy systems with limited samples, showing how these techniques can be adapted to different industrial contexts. These applications highlight how ML-based fault diagnosis is becoming essential to competitive industrial operations.

The demand for interpretable models in industrial settings has led to important advances in explainable fault diagnosis. SHAP (SHapley Additive exPlanations) \citep{lundberg2017} has emerged as a valuable tool for identifying influential variables in industrial fault classification, providing operators with insights that guide troubleshooting efforts. For instance, \citet{lemos2023} applied SHAP for feature selection on the TEP dataset, enhancing fault detection performance with a reduced feature set. \citet{obanya2024} introduced a permutation entropy method for variable contribution analysis in the TEP process, helping identify fault-responsible variables independently of detection statistics—a capability that reduces diagnosis time in industrial environments. The importance of transparent decision-making in critical industrial scenarios was emphasized by \citet{maged2024} in their review of explainable AI for fault detection. Advanced architectures like the deep residual shrinkage networks proposed by \citet{zhao2020} enhance model interpretability while maintaining diagnostic performance, addressing the industrial requirement for both accuracy and explainability.

Causal analysis offers deeper insights into fault mechanisms than correlation-based approaches alone. Causal discovery algorithms—PC \citep{spirtes2001}, FCI \citep{spirtes2001}, RFCI \citep{colombo2012}, LINGAM \citep{shimizu2006}, and NOTEARS \citep{zheng2018}—construct Directed Acyclic Graphs (DAGs) that represent causal relationships between industrial process variables. \citet{zheng2025} extended this approach with a causal graph-based spatial-temporal attention network for remaining useful life prediction in complex industrial systems. These DAGs help identify root causes of industrial faults, reduce false positives, and improve diagnosis model robustness by incorporating domain knowledge—all critical benefits in production environments where rapid, accurate diagnosis directly impacts operational efficiency.

The integration of causal analysis with machine learning represents a promising direction for industrial fault detection. \citet{liu2024} developed a graph attention network with Granger causality mapping specifically for fault detection and root cause diagnosis in the TEP dataset, demonstrating how causal maps can pinpoint the sources of process deviations in industrial systems. This approach aligns with Industry 4.0 initiatives, where data fusion and machine learning increasingly drive industrial prognosis, as outlined by \citet{diez-olivan2019}.

Despite these advancements, industrial applications have seen limited research integrating SHAP-based feature selection with causal analysis for fault classification—a combination that could significantly improve both accuracy and interpretability. Most existing studies focus on either interpretability (e.g., SHAP) or causality (e.g., DAGs) in isolation, rather than combining these approaches to leverage their complementary strengths. This work bridges this gap by integrating SHAP for feature selection with multiple causal analysis algorithms (PC, FCI, RFCI, LINGAM, and NOTEARS) to construct DAGs, providing industrial engineers with a comprehensive framework for fault diagnosis that delivers both accurate detection and causal insights for effective remediation strategies.

\section{Material and Methods}
\label{sec:background}

Fault detection in industrial systems is increasingly critical as processes grow in complexity and generate vast amounts of data, posing significant challenges for ensuring operational safety, efficiency, and reliability. Traditional fault detection methods often struggle with high-dimensional datasets and fail to provide interpretable insights into the causal mechanisms driving faults, leading to delayed responses and costly downtime in industries like chemical manufacturing. This section lays the foundation for our study by introducing a novel framework that integrates SHAP-based feature selection with causal analysis to enhance fault detection in industrial processes. We first define the problem using the Tennessee Eastman Process (TEP), a widely recognized benchmark for evaluating fault detection strategies in chemical engineering. We then describe our integrated fault detection framework, which combines machine learning, feature selection, and causal inference to achieve accurate fault detection while offering actionable insights into fault propagation and root causes, addressing both practical and theoretical needs in industrial engineering.

\subsection{Problem Description}

The Tennessee Eastman Process (TEP) is a widely used benchmark for evaluating fault detection methods in industrial settings. Proposed by \citet{downs1993}, the TEP simulates a realistic industrial process involving a series of chemical reactions, separation, and recycling operations. The process consists of five major units: a reactor, condenser, compressor, vapor-liquid separator, and stripper, as illustrated in the detailed schematic diagram in Fig. 1. A simplified process flow diagram of the TEP, highlighting the main units and material flows, is provided in Fig. 2 for clarity.

\begin{figure}[!htbp]
\centering
\includegraphics[width=0.9\textwidth]{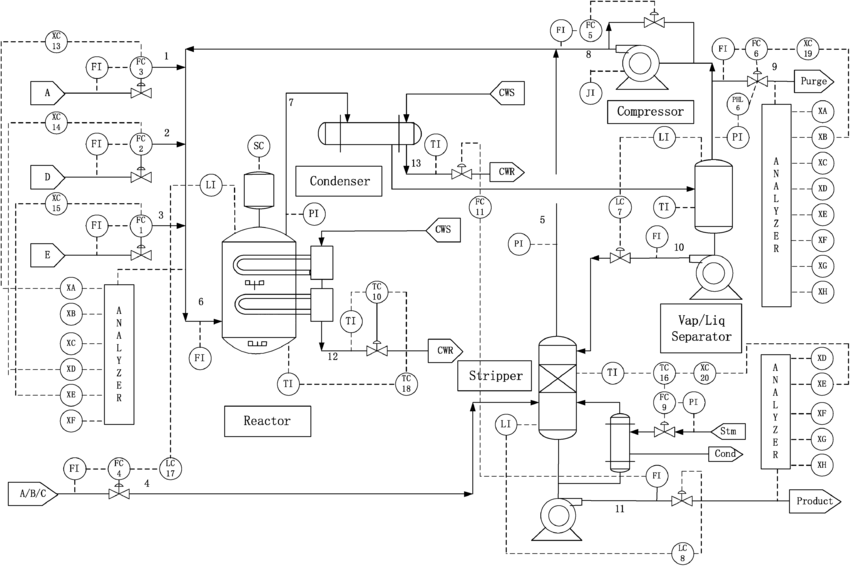}
\caption{Tennessee Eastman Process (TEP) diagram \citep{li2011}.}
\label{fig:tep_li}
\end{figure}

\begin{figure}[!htbp]
\centering
\includegraphics[width=0.9\textwidth]{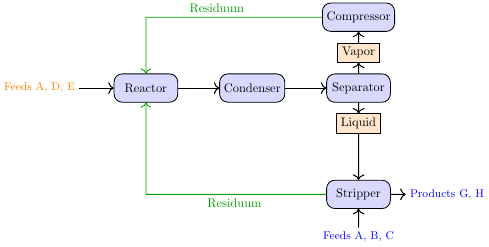}
\caption{Simplified process flow diagram of TEP.}
\label{fig:tep_simplified}
\end{figure}

The TEP involves eight components (A, B, C, D, E, F, G, and H), with feeds A, D, and E entering the reactor, where they undergo chemical reactions to produce products G and H, along with byproducts such as F. The reactor output is cooled in the condenser, and the resulting vapor and liquid streams are separated in the vapor-liquid separator. The vapor stream is recycled back to the reactor via the compressor, while the liquid stream is sent to the stripper, where additional feeds A, B, and C are introduced to remove unreacted components. The final products G and H are obtained from the stripper, with residuum streams recycled back to the reactor. This process is characterized by complex interactions between variables, making it an ideal testbed for fault detection studies.

Table 1 lists the manipulated variables, including feed flow rates, reactor cooling water flow, and agitator speed, along with their base case values and operational limits. Table 2 lists the continuous process measurements, such as reactor temperature, pressure, and product separator levels, which are critical for monitoring the process. Additionally, Table 3 presents the sampled process measurements, including the composition of the reactor feed, purge gas, and product streams, which are analyzed at different sampling frequencies. These variables are subject to disturbances, as outlined in Table 4, which describes 20 independent disturbance variables (IDVs) that can affect the process, ranging from step changes in feed composition to random variations in reactor kinetics.

\begin{table}[!htbp]
    \centering
    \caption{Process manipulated variables}
    \label{tab:tep-manipulated-variables}
    \small
    \begin{tabular}{lccccc}
        \hline
        \textbf{Variable name} & \textbf{Number} & \textbf{Base (\%)} & \textbf{Low} & \textbf{High} & \textbf{Units} \\
        \hline
        D feed flow (stream 2) & XMV (1) & 63.053 & 0 & 5811 & kg h$^{-1}$ \\
        E feed flow (stream 3) & XMV (2) & 53.980 & 0 & 8354 & kg h$^{-1}$ \\
        A feed flow (stream 1) & XMV (3) & 24.644 & 0 & 1.017 & kscmh \\
        A and C feed flow (stream 4) & XMV (4) & 61.302 & 0 & 15.25 & kscmh \\
        Compressor recycle valve & XMV (5) & 22.210 & 0 & 100 & \% \\
        Purge valve (stream 9) & XMV (6) & 40.064 & 0 & 100 & \% \\
        Separator pot liquid flow (stream 10) & XMV (7) & 38.100 & 0 & 65.71 & m$^3$ h$^{-1}$ \\
        Stripper liquid product flow (stream 11) & XMV (8) & 46.534 & 0 & 49.10 & m$^3$ h$^{-1}$ \\
        Stripper steam valve & XMV (9) & 47.446 & 0 & 100 & \% \\
        Reactor cooling water flow & XMV (10) & 41.106 & 0 & 227.1 & m$^3$ h$^{-1}$ \\
        Condenser cooling water flow & XMV (11) & 18.114 & 0 & 272.6 & m$^3$ h$^{-1}$ \\
        \hline
    \end{tabular}
\end{table}

\begin{table}[!htbp]
    \centering
    \caption{Continuous process measurements}
    \label{tab:tep-continuous-measurements}
    \small
    \begin{tabular}{lccc}
        \hline
        \textbf{Variable name} & \textbf{Number} & \textbf{Base value} & \textbf{Units} \\
        \hline
        A feed (stream 1) & XMEAS (1) & 0.25052 & kscmh \\
        D feed (stream 2) & XMEAS (2) & 3664.0 & kg h$^{-1}$ \\
        E feed (stream 3) & XMEAS (3) & 4509.3 & kg h$^{-1}$ \\
        A and C feed (stream 4) & XMEAS (4) & 9.3477 & kscmh \\
        Recycle flow (stream 8) & XMEAS (5) & 26.902 & kscmh \\
        Reactor feed rate (stream 6) & XMEAS (6) & 42.339 & kscmh \\
        Reactor pressure & XMEAS (7) & 2705.0 & kPa gauge \\
        Reactor level & XMEAS (8) & 75.000 & \% \\
        Reactor temperature & XMEAS (9) & 120.40 & $^\circ$C \\
        Purge rate (stream 9) & XMEAS (10) & 0.33712 & kscmh \\
        Product separator temperature & XMEAS (11) & 80.109 & $^\circ$C \\
        Product separator level & XMEAS (12) & 50.000 & \% \\
        Product separator pressure & XMEAS (13) & 2633.7 & kPa gauge \\
        Product separator underflow (stream 10) & XMEAS (14) & 25.160 & m$^3$ h$^{-1}$ \\
        Stripper level & XMEAS (15) & 50.000 & \% \\
        Stripper pressure & XMEAS (16) & 3102.2 & kPa gauge \\
        Stripper underflow (stream 11) & XMEAS (17) & 22.949 & m$^3$ h$^{-1}$ \\
        Stripper temperature & XMEAS (18) & 65.731 & $^\circ$C \\
        Stripper steam flow & XMEAS (19) & 230.31 & kg h$^{-1}$ \\
        Compressor work & XMEAS (20) & 341.43 & kW \\
        Reactor cooling water outlet temp. & XMEAS (21) & 94.599 & $^\circ$C \\
        Separator cooling water outlet temp. & XMEAS (22) & 77.297 & $^\circ$C \\
        \hline
    \end{tabular}
\end{table}

\begin{table}[!htbp]
    \centering
    \caption{Sampled process measurements}
    \label{tab:tep-sampled-measurements}
    \small
    \begin{tabular}{lcccc}
        \hline
        \textbf{Variable} & \textbf{Number} & \textbf{Base value} & \textbf{Units} & \textbf{Sampling / Dead} \\
        \hline
        \multicolumn{5}{c}{\textbf{Reactor feed analysis (stream 6)}} \\ \hline
        A & XMEAS (23) & 32.188 & mol\% & 0.1 h / 0.1 h \\
        B & XMEAS (24) & 8.8933 & mol\% & 0.1 h / 0.1 h \\
        C & XMEAS (25) & 26.383 & mol\% & 0.1 h / 0.1 h \\
        D & XMEAS (26) & 6.8820 & mol\% & 0.1 h / 0.1 h \\
        E & XMEAS (27) & 18.776 & mol\% & 0.1 h / 0.1 h \\
        F & XMEAS (28) & 1.6567 & mol\% & 0.1 h / 0.1 h \\ \hline
        \multicolumn{5}{c}{\textbf{Purge gas analysis (stream 9)}} \\ \hline
        A & XMEAS (29) & 32.958 & mol\% & 0.1 h / 0.1 h \\
        B & XMEAS (30) & 13.823 & mol\% & 0.1 h / 0.1 h \\
        C & XMEAS (31) & 23.978 & mol\% & 0.1 h / 0.1 h \\
        D & XMEAS (32) & 1.2565 & mol\% & 0.1 h / 0.1 h \\
        E & XMEAS (33) & 18.379 & mol\% & 0.1 h / 0.1 h \\
        F & XMEAS (34) & 2.2633 & mol\% & 0.1 h / 0.1 h \\
        G & XMEAS (35) & 4.8436 & mol\% & 0.1 h / 0.1 h \\
        H & XMEAS (36) & 2.2996 & mol\% & 0.1 h / 0.1 h \\ \hline
        \multicolumn{5}{c}{\textbf{Product analysis (stream 11)}} \\ \hline
        D & XMEAS (37) & 0.017877 & mol\% & 0.25 h / 0.25 h \\
        E & XMEAS (38) & 0.833570 & mol\% & 0.25 h / 0.25 h \\
        F & XMEAS (39) & 0.099585 & mol\% & 0.25 h / 0.25 h \\
        G & XMEAS (40) & 53.724 & mol\% & 0.25 h / 0.25 h \\
        H & XMEAS (41) & 43.828 & mol\% & 0.25 h / 0.25 h \\
        \hline
    \end{tabular}
\end{table}

\begin{table}[!htbp]
    \centering
    \caption{Process disturbances}
    \label{tab:tep-disturbances}
    \small
    \begin{tabular}{lcc}
        \hline
        \textbf{Variable number} & \textbf{Variable} & \textbf{Type} \\
        \hline
        IDV (1) & A/C feed ratio, B composition constant (stream 4) & Step \\
        IDV (2) & B composition, A/C ratio constant (stream 4) & Step \\
        IDV (3) & D feed temperature (stream 2) & Step \\
        IDV (4) & Reactor cooling water inlet temperature & Step \\
        IDV (5) & Condenser cooling water inlet temperature & Step \\
        IDV (6) & A feed loss (stream 1) & Step \\
        IDV (7) & C header pressure loss (stream 4) & Step \\
        IDV (8) & A, B, C feed composition (stream 4) & Random variation \\
        IDV (9) & D feed temperature (stream 2) & Random variation \\
        IDV (10) & C feed temperature (stream 4) & Random variation \\
        IDV (11) & Reactor cooling water inlet temperature & Random variation \\
        IDV (12) & Condenser cooling water inlet temperature & Random variation \\
        IDV (13) & Reaction kinetics & Slow drift \\
        IDV (14) & Reactor cooling water valve & Sticking \\
        IDV (15) & Condenser cooling water valve & Sticking \\
        IDV (16) & Unknown & Unknown \\
        IDV (17) & Unknown & Unknown \\
        IDV (18) & Unknown & Unknown \\
        IDV (19) & Unknown & Unknown \\
        IDV (20) & Unknown & Unknown \\
        \hline
    \end{tabular}
\end{table}

The TEP is designed to operate under a base case scenario, where the process variables are maintained at their nominal values, however, the introduction of disturbances (IDVs) can lead to faults, such as deviations in reactor temperature or product composition, which must be detected and mitigated to ensure safe and efficient operation. The complexity of the TEP, combined with its realistic simulation of industrial challenges, makes it an ideal platform for testing advanced fault detection methods.

\clearpage

\subsection{Machine Learning Models}

We evaluated three machine learning algorithms commonly used in industrial applications, specifically selected to address different aspects of fault detection in chemical processes.

\subsubsection{Multilayer Perceptron (MLP)}

The Multilayer Perceptron (MLP) \citep{rumelhart1986} is a feedforward neural network designed to model complex, nonlinear relationships in data. It consists of interconnected layers of nodes, where input data passes through hidden layers to produce an output. Each node processes weighted inputs using an activation function, enabling the network to learn intricate patterns. In fault detection, MLP excels at capturing subtle variations in high-dimensional industrial data, making it suitable for identifying anomalies in processes like the TEP. Its flexibility allows adaptation to diverse fault patterns, though careful tuning is needed to ensure robust performance.

\subsubsection{XGBoost (XGB)}

XGBoost \citep{chen2016} is a gradient boosting algorithm that builds an ensemble of decision trees to achieve high predictive accuracy. It iteratively constructs trees, with each tree correcting errors from the previous ones by optimizing a loss function. XGBoost incorporates regularization to prevent overfitting and handles imbalanced datasets effectively, which is critical for industrial fault detection. Its ability to capture complex variable interactions makes it well-suited for identifying faults in the TEP, where process dynamics are nonlinear. The algorithm's efficiency and robustness enhance its applicability in real-time monitoring scenarios.

\subsubsection{k-Nearest Neighbors (KNN)}

The k-Nearest Neighbors (KNN) algorithm \citep{cover1967} is a non-parametric method that classifies data points based on the majority class of their $k$ closest neighbors in the feature space. It calculates distances to identify nearby instances, making it intuitive for detecting localized patterns. In fault detection, KNN excels at identifying anomalies that form distinct clusters, such as specific fault conditions in the TEP. Its simplicity ensures interpretable results, valuable for process operators, though performance depends on choosing an appropriate distance metric and $k$ value.

\subsection{Data Preprocessing and Imbalance Handling}

Industrial fault datasets typically suffer from significant class imbalance, with normal operations far outnumbering fault conditions. To address this challenge, we implemented a two-stage resampling approach combining synthetic oversampling and targeted undersampling techniques. This resampling was applied only to training data, ensuring test sets maintained original distributions for realistic evaluation.

\subsubsection{SMOTE Oversampling}

Synthetic Minority Over-sampling Technique (SMOTE) \citep{chawla2002} addresses the minority class imbalance by creating synthetic samples rather than simple duplication. The algorithm identifies k-nearest neighbors for minority class instances and generates new samples along the lines connecting these instances in the feature space. This approach preserves the complex patterns and distributions present in fault scenarios while expanding the available training examples.

\subsubsection{Random Undersampling}

Following the oversampling stage, we employed random undersampling \citep{japkowicz2002} to reduce redundancy in the majority class. This technique randomly removes samples from the over-represented normal operation class, creating a more balanced overall class distribution. The approach helps prevent classifier bias toward the majority class while maintaining sufficient examples of normal operation patterns.

The combination of SMOTE and random undersampling creates a training dataset that better represents the full spectrum of operational states, while preserving original data characteristics—essential for industrial systems where maintaining fidelity to real process behavior is critical.

\subsection{SHAP-based Feature Selection}

To address the high dimensionality of industrial data, we employed SHAP (SHapley Additive exPlanations) for feature selection, as proposed by \citet{lundberg2017}. SHAP values, derived from game-theoretic principles, provide consistent and interpretable feature importance metrics by quantifying each variable's contribution to model predictions, accounting for feature interactions, and offering both global (dataset-wide) and local (instance-specific) explanations. This approach is particularly valuable in complex industrial processes, where understanding feature interactions is critical for reliable fault detection.

\subsection{Causal Analysis with Directed Acyclic Graphs}

To enhance interpretability and provide actionable insights for process operators, we applied five most known causal discovery algorithms to construct Directed Acyclic Graphs (DAGs) from the selected feature subsets. These algorithms reveal the underlying causal structure of process variables, enabling a deeper understanding of fault propagation mechanisms.

\subsubsection{PC Algorithm}

The PC algorithm (named after its inventors, Peter and Clark) \citep{spirtes2001} is a constraint-based method that begins with a fully connected undirected graph and systematically removes edges based on conditional independence tests. The algorithm assumes causal sufficiency (no unmeasured common causes) and faithfulness (all conditional independencies in the data reflect the underlying causal structure).

In industrial settings, the PC algorithm provides an efficient approach for initial causal structure exploration, identifying direct causal influences between process variables.

\subsubsection{Fast Causal Inference (FCI)}

The Fast Causal Inference algorithm \citep{spirtes2001} extends the PC algorithm to handle latent confounders (unmeasured common causes), a common situation in industrial settings where not all relevant process variables can be measured. FCI produces a Partial Ancestral Graph (PAG) that represents both direct causal relationships and potential hidden confounders.

\subsubsection{Really Fast Causal Inference (RFCI)}

The Really Fast Causal Inference algorithm \citep{colombo2012} offers a computationally efficient alternative to FCI, particularly valuable for high-dimensional industrial datasets. RFCI approximates the output of FCI while requiring fewer conditional independence tests, making it practical for large-scale industrial systems with many process variables.

\subsubsection{Linear Non-Gaussian Acyclic Model (LINGAM)}

LINGAM \citep{shimizu2006} takes a different approach to causal discovery, assuming linear relationships with non-Gaussian noise between variables. This distinctive approach allows LINGAM to identify the direction of causality even in cases where constraint-based methods like PC would find multiple equivalent causal structures. The algorithm uses Independent Component Analysis (ICA) to estimate the causal structure, providing quantitative estimates of causal effect strength between process variables.

\subsubsection{NOTEARS}

NOTEARS \citep{zheng2018} represents a significant methodological advancement in causal discovery, formulating DAG learning as a continuous optimization problem rather than a combinatorial search. This approach allows NOTEARS to scale efficiently to high-dimensional industrial datasets while providing weighted edges that indicate the strength of causal relationships.

These five causal discovery algorithms provide complementary perspectives on the causal structure of industrial processes. By applying multiple algorithms and comparing their outputs, we can identify robust causal relationships that are consistently detected across different methodological approaches, providing more reliable insights into fault propagation pathways.

\section{Experimental Methodology}
\label{sec:exp_methgy}

To evaluate our approach for industrial fault detection, we designed a comprehensive assessment methodology that ensures both statistical rigor and industrial relevance. The framework encompasses cross-validation procedures, multiple performance metrics, and a structured experimental workflow tailored to the challenges of industrial settings.

\subsection{Integrated Fault Detection Framework}

The framework consists of four interconnected components that work synergistically to deliver both high detection accuracy and actionable process insights: (1) A multi-model machine learning approach that evaluates and optimizes various algorithms specifically selected for their suitability to industrial time-series data; (2) A data preprocessing pipeline that handles the inherent class imbalance in industrial fault datasets while preserving the integrity of the original process characteristics; (3) A SHAP-based feature selection methodology that systematically reduces dimensionality while maintaining or improving detection performance, creating a more manageable and interpretable variable space; and (4) A comprehensive causal analysis component that employs multiple DAG causal discovery algorithms to identify the underlying causal relationships between process variables and fault conditions.

\begin{figure}[!htbp]
\centering
\includegraphics[width=\textwidth]{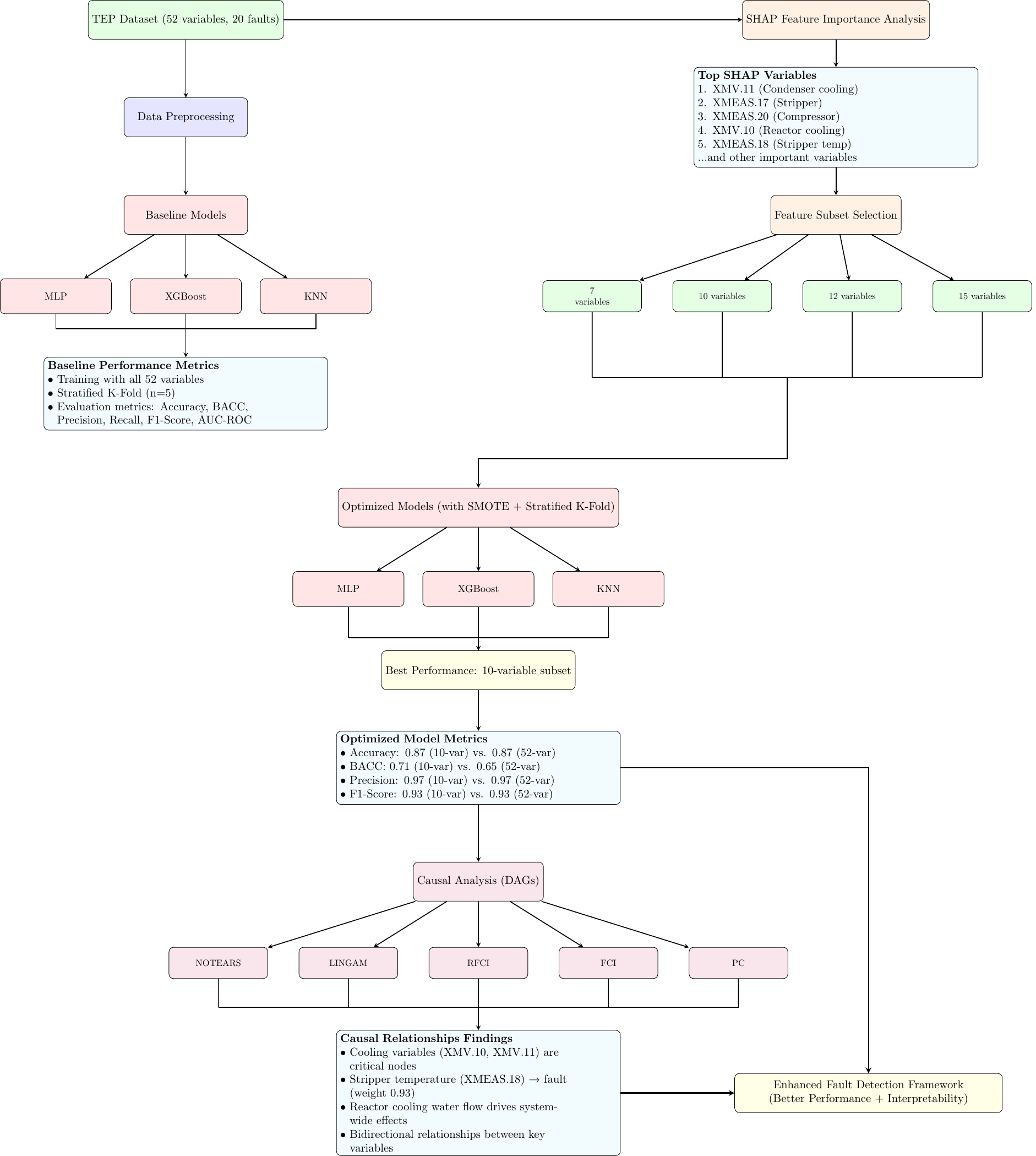}
\caption{Framework integrating SHAP-based feature selection with causal analysis for industrial fault detection. The workflow shows how dimensionality reduction to just 10 variables maintains model performance while enabling interpretable causal analysis through multiple DAG algorithms.}
\label{fig:tep-workflow}
\end{figure}

Our experimental workflow consisted of five stages: (1) training baseline models with all variables, (2) ranking variables by SHAP importance, (3) evaluating models with reduced variable subsets, (4) applying causal discovery algorithms to identify relationship structures, and (5) analyzing the resulting DAGs to identify consistent causal patterns relevant to industrial operations.

\subsection{Cross-Validation}

To ensure robust and reliable evaluation, we implemented stratified 5-fold cross-validation on our training dataset, while maintaining a separate test dataset for final performance assessment. Stratification ensures that each fold maintains approximately the same class distribution as the full dataset, which is particularly important for imbalanced datasets like TEP. The choice of $k=5$ strikes a balance between computational cost and estimation reliability, providing sufficient samples in each fold while allowing for appropriate variance estimates.

\subsection{Performance Metrics}

To evaluate the performance of the classification algorithms applied in this context, six widely used metrics relevant to industrial fault detection from the literature were employed: Accuracy (ACC), Balanced Accuracy (BACC), Precision, Recall, F1-Score and AUC-ROC. The equations for these metrics are as follows:

\begin{equation}
\text{ACC} = \frac{TP + TN}{TP + FP + TN + FN}
\end{equation}

\begin{equation}
\text{BACC} = \frac{\frac{TP}{TP + FN} + \frac{TN}{TN + FP}}{2}
\end{equation}

\begin{equation}
\text{Recall} = \frac{TP}{TP + FN}
\end{equation}

\begin{equation}
\text{Precision} = \frac{TP}{TP + FP}
\end{equation}

\begin{equation}
\text{F1-Score} = 2 \times \frac{\text{Recall} \times \text{Precision}}{\text{Recall} + \text{Precision}}
\end{equation}

\begin{equation}
\text{AUC-ROC} = \int_{0}^{1} TPR(FPR^{-1}(t)) \, dt
\end{equation}

The variables $TP$, $TN$, $FP$, and $FN$ refer, respectively, to the values of true positives, true negatives, false positives, and false negatives. The variable $TPR$ is the True Positive Rate (Recall), $FPR$ is the False Positive Rate, and $FPR^{-1}$ is the inverse function of $FPR$. This approach provides a comprehensive assessment of model performance across different operational priorities. In industrial settings, balanced accuracy is particularly important as it measures the ability to detect rare faults—critical for processes where missed faults can have severe safety or economic consequences.

\subsection{Implementation Details}
\label{subsec-hyperparameter-tuning}

The performance of machine learning models for fault detection depends significantly on their hyperparameter configurations. To ensure optimal model performance while avoiding manual trial-and-error, we implemented RandomizedSearchCV \citep{bergstra2012} with a carefully designed search space for each classifier. The hyperparameter optimization process used 5-fold cross-validation with the F1-macro score as the optimization metric, chosen specifically to account for the class imbalance in the TEP dataset. Each RandomizedSearchCV was configured to perform 20 iterations, striking a balance between exploration of the hyperparameter space and computational efficiency. Tables 5-7 present the optimal hyperparameter configurations identified for each model variant.

\begin{table}[!htbp]
    \centering
    \caption{Optimal hyperparameter configurations for MLP models}
    \label{tab:mlp-hyperparameters}
    \small
    \begin{tabular}{lccccc}
        \hline
        \textbf{Vars} & \textbf{Hidden Layers} & \textbf{Activation} & \textbf{Learn Rate} & \textbf{Reg.} & \textbf{Batch} \\
        \hline
        52 & (300, 150, 75) & relu & 0.001 & 0.0001 & 128 \\
        7 & (300, 150, 75) & relu & 0.001 & 0.001 & 128 \\
        10 & (250, 125, 60) & relu & 0.001 & 0.0001 & 64 \\
        12 & (250, 125, 60) & relu & 0.001 & 0.001 & 64 \\
        15 & (250, 125, 60) & relu & 0.005 & 0.0001 & 128 \\
        \hline
    \end{tabular}
\end{table}

\begin{table}[!htbp]
    \centering
    \caption{Optimal hyperparameter configurations for XGBoost models}
    \label{tab:xgboost-hyperparameters}
    \small
    \begin{tabular}{lcccccc}
        \hline
        \textbf{Vars} & \textbf{Learn} & \textbf{Depth} & \textbf{Est.} & \textbf{Subsamp} & \textbf{Min Child} & \textbf{Gamma} \\
        \hline
        52 & 0.05 & 3 & 200 & 0.9 & 5 & 0 \\
        7 & 0.05 & 3 & 200 & 0.9 & 3 & 0.2 \\
        10 & 0.05 & 3 & 200 & 0.9 & 3 & 0.2 \\
        12 & 0.05 & 3 & 200 & 0.9 & 3 & 0.2 \\
        15 & 0.05 & 3 & 200 & 0.9 & 5 & 0 \\
        \hline
    \end{tabular}
\end{table}

\begin{table}[!htbp]
    \centering
    \caption{Optimal hyperparameter configurations for KNN models}
    \label{tab:knn-hyperparameters}
    \small
    \begin{tabular}{lccccc}
        \hline
        \textbf{Variables} & \textbf{$k$} & \textbf{Weights} & \textbf{Metric} & \textbf{$p$} & \textbf{Leaf Size} \\
        \hline
        52 & 5 & uniform & manhattan & 2 & 20 \\
        7 & 5 & uniform & manhattan & 2 & 20 \\
        10 & 5 & uniform & manhattan & 2 & 20 \\
        12 & 5 & uniform & manhattan & 2 & 20 \\
        15 & 5 & uniform & manhattan & 2 & 20 \\
        \hline
    \end{tabular}
\end{table}

\section{Results}
\label{sec:results}

This section presents the experimental results of our integrated framework for fault detection in the Tennessee Eastman Process (TEP) dataset. We first establish a baseline performance utilizing the complete set of 52 variables, followed by an evaluation of SHAP-based feature selection effects using smaller variable subsets. The top overall results are highlighted in \textbf{bold}, while the best outcomes for each algorithm are indicated with an \underline{underline}. Finally, we present a comparative analysis of the causal structures identified by different DAG algorithms, providing insights for industrial process monitoring.

\subsection{Model Performance with Full Dataset (52 Variables)}

To establish a baseline, we evaluated the performance of MLP, XGBoost, and KNN using all 52 variables in the TEP dataset. Table \ref{tab:full-dataset-metrics} summarizes the results across multiple performance metrics.

\begin{table}[!htbp]
    \centering
    \caption{Performance metrics for MLP, XGBoost, and KNN using the full TEP dataset (52 variables)}
    \label{tab:full-dataset-metrics}
    \footnotesize
    \begin{tabular}{lcccccc}
        \toprule
        \textbf{Alg.} & \textbf{Acc.} & \textbf{BACC} & \textbf{Prec.} & \textbf{Recall} & \textbf{F1} & \textbf{AUC} \\
        \midrule
        MLP & \textbf{0.872±0.007} & \textbf{0.650±0.019} & \textbf{0.968±0.002} & \textbf{0.895±0.009} & \textbf{0.930±0.004} & \textbf{0.858±0.007} \\
        XGBoost & 0.806±0.009 & 0.684±0.014 & 0.973±0.002 & 0.819±0.011 & 0.889±0.006 & 0.825±0.003 \\
        KNN & 0.749±0.006 & 0.522±0.010 & 0.955±0.001 & 0.773±0.008 & 0.854±0.004 & 0.607±0.005 \\
        \bottomrule
    \end{tabular}
\end{table}

MLP achieved the highest overall performance with an accuracy of 0.872±0.007 and F1-score of 0.930±0.004, demonstrating its ability to capture complex patterns in the high-dimensional TEP data. However, its balanced accuracy (0.650±0.019) suggests challenges with class imbalance—a common issue in industrial fault detection where normal samples typically outnumber fault instances.

XGBoost showed the highest balanced accuracy (0.684±0.014), indicating better performance across all fault classes despite lower overall accuracy. This makes XGBoost particularly valuable in industrial settings where detecting minority fault classes is critical.

KNN produced the weakest results across most metrics, particularly in balanced accuracy (0.522±0.010) and AUC-ROC (0.607±0.005), highlighting its limitations with high-dimensional, imbalanced industrial process data.

\subsection{SHAP-Based Feature Selection Results}

\begin{figure}[!htbp]
\centering
\includegraphics[width=0.9\textwidth]{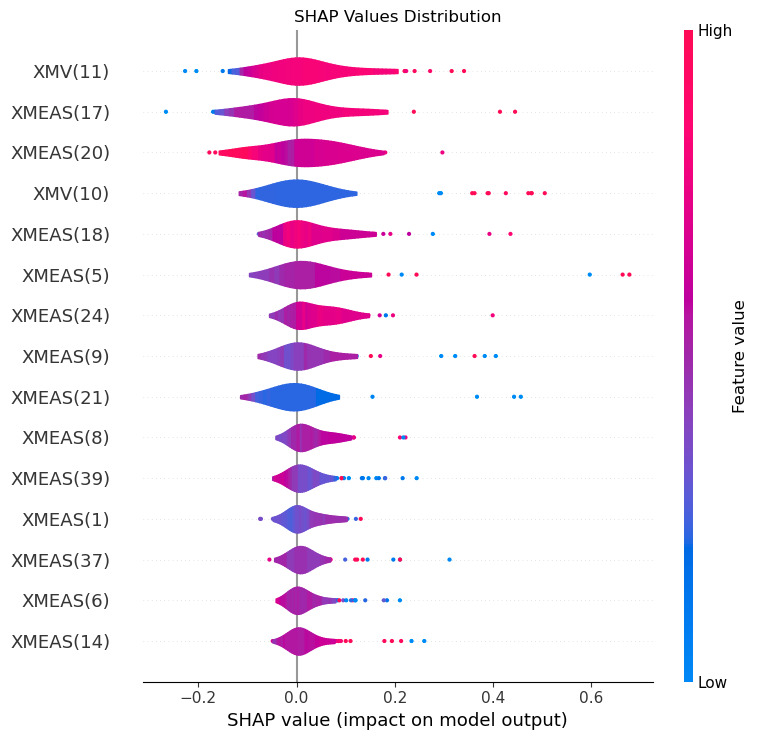}
\caption{SHAP values for the top 15 variables in the TEP dataset. Variables such as XMV.11 (Condenser cooling water flow) and XMEAS.17 (Stripper underflow) have the highest impact on model predictions. Higher SHAP values indicate greater importance for fault detection.}
\label{fig:shap-values}
\end{figure}

To address the high dimensionality of the TEP dataset, we applied SHAP analysis to identify the most influential variables for fault detection. Fig. \ref{fig:shap-values} shows the top 15 variables ranked by their SHAP values, revealing key process parameters that contribute most significantly to model predictions.

The SHAP analysis revealed that cooling-related variables (XMV.11, XMV.10) and stripper parameters (XMEAS.17, XMEAS.18) are particularly influential, aligning with process engineering knowledge about the critical role of thermal management and separation processes in industrial plant stability. From an industrial engineering perspective, this identification of critical variables enables more targeted monitoring and maintenance strategies. Using these SHAP importance rankings, we constructed four variable subsets containing the top 7, 10, 12, and 15 variables to systematically evaluate the impact of dimensionality reduction on model performance.

\subsection{Model Performance with Dimensionality Reduction}

Using the SHAP-selected variable subsets, we retrained our machine learning models and evaluated their performance. Table 8 presents the results for each algorithm.

\begin{table}[!htbp]
    \caption{Performance metrics for MLP, XGBoost, and KNN using SHAP-selected subsets}
    \label{tab:reduced-dataset-metrics}
    \centering
    \footnotesize
    \setlength{\tabcolsep}{3pt}
    \begin{tabular}{llcccccc}
        \toprule
        \textbf{Alg.} & \textbf{V} & \textbf{Acc.} & \textbf{BACC} & \textbf{Prec.} & \textbf{Rec.} & \textbf{F1} & \textbf{AUC} \\
        \midrule
        \multirow{4}{*}{MLP}
        & 7  & 0.842±0.016 & 0.696±0.033 & 0.974±0.004 & 0.858±0.021 & 0.912±0.010 & 0.850±0.005 \\
        & 10 & \textbf{0.872±0.006} & \textbf{0.711±0.027} & \textbf{0.974±0.003} & \textbf{0.889±0.009} & \textbf{0.930±0.004} & \textbf{0.869±0.002} \\
        & 12 & 0.868±0.007 & 0.712±0.030 & 0.975±0.004 & 0.884±0.010 & 0.927±0.004 & 0.867±0.003 \\
        & 15 & 0.871±0.007 & 0.670±0.026 & 0.970±0.003 & 0.892±0.007 & 0.929±0.004 & 0.860±0.003 \\
        \midrule
        \multirow{4}{*}{XGB}
        & 7  & 0.759±0.008 & 0.716±0.008 & 0.979±0.001 & 0.763±0.009 & 0.857±0.006 & 0.809±0.004 \\
        & 10 & 0.780±0.006 & 0.719±0.008 & 0.978±0.001 & 0.786±0.007 & 0.872±0.004 & 0.823±0.002 \\
        & 12 & 0.786±0.005 & 0.703±0.012 & 0.976±0.002 & 0.795±0.007 & 0.876±0.004 & 0.820±0.004 \\
        & 15 & \underline{0.790±0.005} & \underline{0.694±0.009} & \underline{0.975±0.001} & \underline{0.800±0.006} & \underline{0.879±0.004} & \underline{0.819±0.003} \\
        \midrule
        \multirow{4}{*}{KNN}
        & 7  & \underline{0.737±0.006} & \underline{0.625±0.014} & \underline{0.968±0.002} & \underline{0.749±0.007} & \underline{0.844±0.004} & \underline{0.689±0.013} \\
        & 10 & 0.734±0.001 & 0.628±0.007 & 0.968±0.001 & 0.745±0.002 & 0.842±0.001 & 0.687±0.010 \\
        & 12 & 0.731±0.007 & 0.602±0.010 & 0.965±0.001 & 0.744±0.008 & 0.840±0.005 & 0.666±0.013 \\
        & 15 & 0.727±0.006 & 0.563±0.003 & 0.960±0.000 & 0.744±0.007 & 0.839±0.004 & 0.625±0.008 \\
        \bottomrule
    \end{tabular}
\end{table}

Using the SHAP-selected variable subsets, we retrained our machine learning models and evaluated their performance. The results demonstrate several key findings with significant implications for industrial fault detection. First, MLP with 10 variables achieved the best overall performance, matching the accuracy of the full 52-variable model (0.872) while improving balanced accuracy from 0.650 to 0.711. This represents an 80\% reduction in input dimensionality with improved performance across all metrics—a substantial efficiency gain for industrial monitoring systems.

Second, balanced accuracy improved for all models with reduced variable sets, particularly for the 7 and 10-variable configurations. XGBoost showed consistent performance across variable subsets, with all configurations maintaining strong balanced accuracy (between 0.694 and 0.719). The 10-variable model provided the best balance between dimensionality reduction and performance with the highest balanced accuracy (0.719) and AUC-ROC (0.823), making it particularly suitable for real-time industrial monitoring applications.

KNN benefited least from dimensionality reduction, showing only marginal improvement in balanced accuracy with the 7 and 10-variable subsets compared to the full model. This aligns with KNN's known sensitivity to the curse of dimensionality—an important consideration for implementation in industrial settings.

These findings validate our SHAP-based feature selection approach, indicating that carefully selected subsets of variables can maintain or improve classification performance while significantly reducing computational complexity and enhancing model interpretability. For industrial applications, this means reduced processing requirements and more focused monitoring of critical variables.

\subsection{Causal Analysis Results}

Based on the performance metrics, we selected the 10-variable subset for causal analysis, applying five different causal discovery algorithms to uncover the underlying relationships between process variables and fault conditions. This subset represents an optimal balance between dimensionality reduction and model performance, particularly for the MLP classifier.

\begin{figure}[!htbp]
\centering
\includegraphics[width=\textwidth]{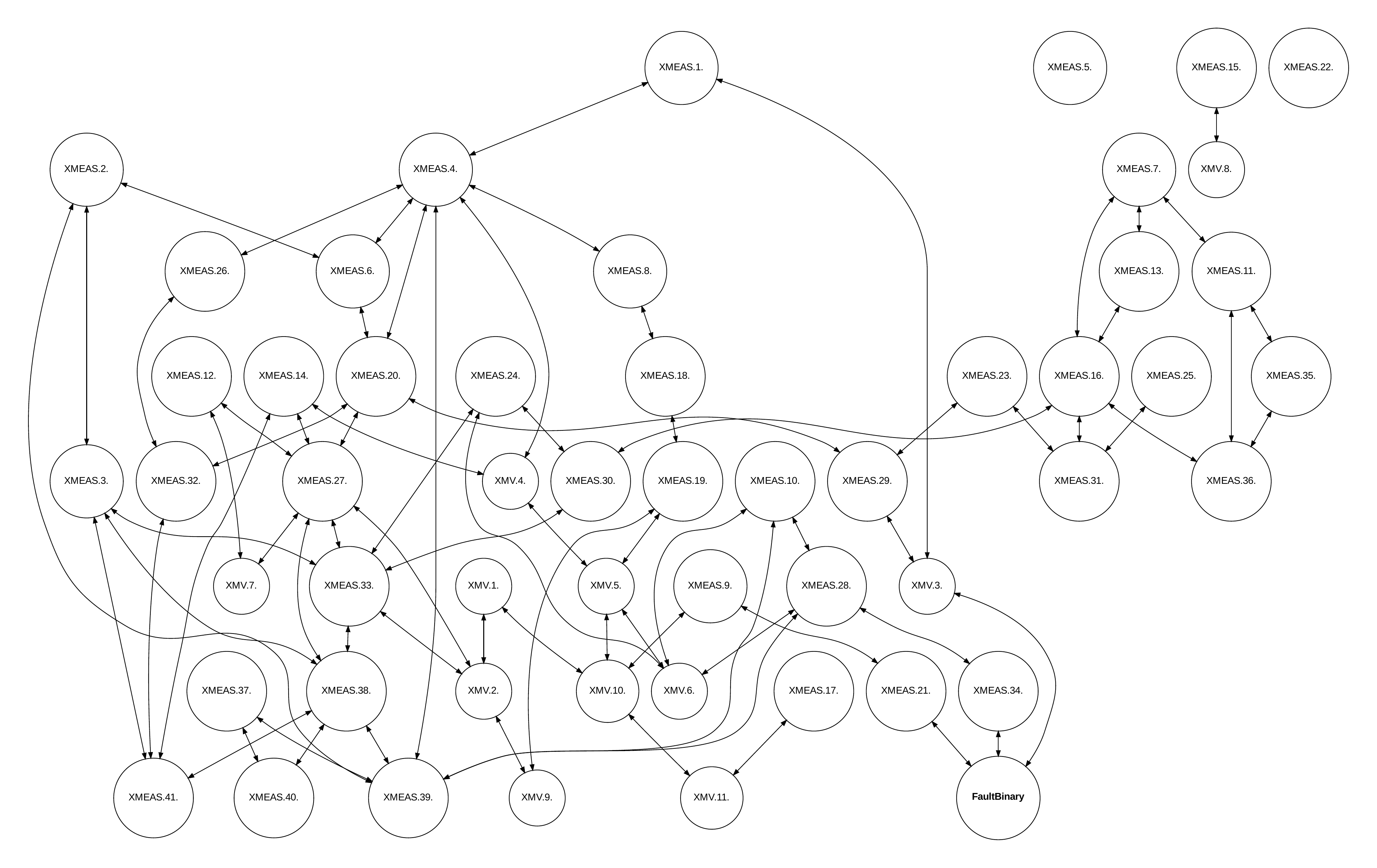}
\caption{Complete 52-variable DAG of the TEP dataset (RFCI algorithm) demonstrating the inherent complexity of industrial process monitoring.}
\label{fig:full-dag}
\end{figure}

To illustrate the motivation for dimensionality reduction in causal analysis, Fig. \ref{fig:full-dag} presents the complete DAG constructed from all 52 TEP variables using the RFCI algorithm. This highly complex network demonstrates the challenges of interpreting causal relationships in high-dimensional industrial data, with numerous interconnections that obscure clear causal pathways.

The causal discovery results from the reduced 10-variable subset complement and extend the insights gained from SHAP-based feature selection, offering a dual lens through which to understand fault detection in the Tennessee Eastman Process. Variables identified as highly important by SHAP—particularly XMV.11 and XMEAS.17—consistently emerged as causally significant across all five algorithms, confirming their pivotal role in fault mechanisms from both predictive and causal perspectives.

\begin{figure}[!ht]
\centering
\includegraphics[width=0.6\textwidth]{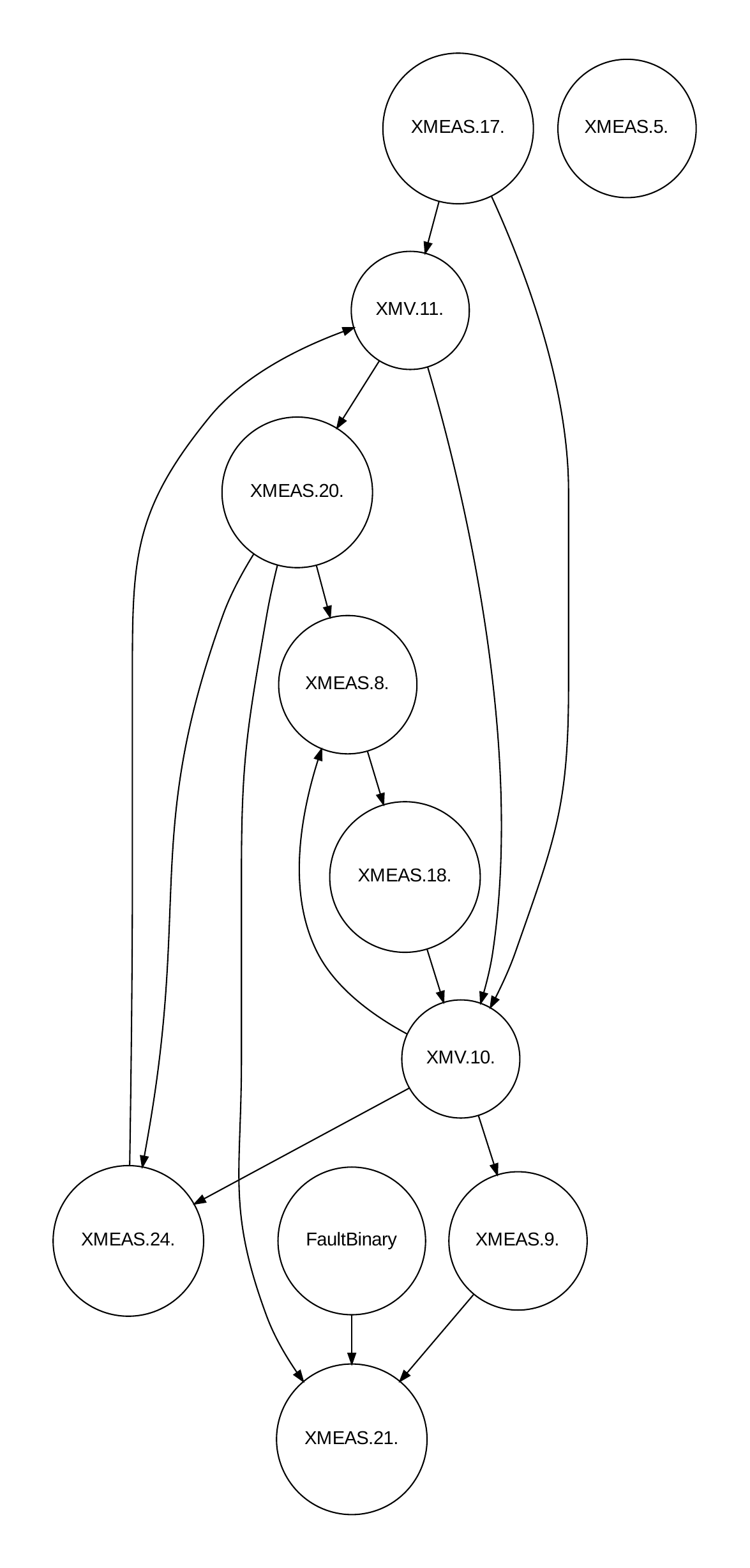}
\caption{DAG generated by the PC algorithm for the 10-variable subset. This structure positions XMEAS.17 (Stripper underflow) as a root node, indicating its fundamental role in initiating causal chains that ultimately affect fault detection through XMEAS.21 (Reactor cooling water outlet temperature).}
\label{fig:dag-pc}
\end{figure}

\begin{figure}[!ht]
\centering
\includegraphics[width=0.65\textwidth]{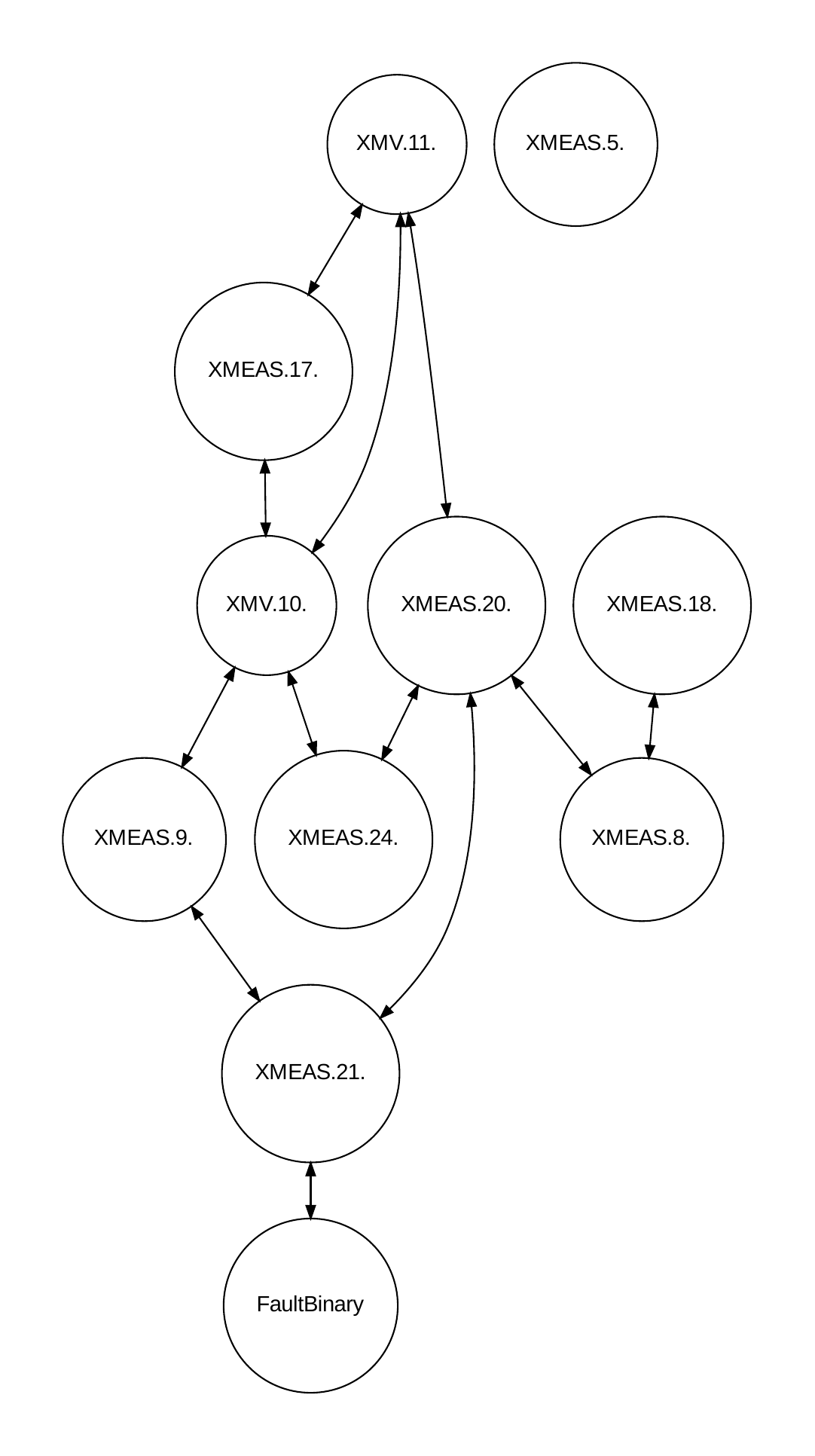}
\caption{DAG generated by the FCI algorithm for the 10-variable subset. This graph highlights bidirectional relationships between XMV.11 (Condenser cooling water flow) and XMEAS.17 (Stripper underflow), capturing the complex control loops characteristic of industrial processes.}
\label{fig:dag-fci}
\end{figure}

\begin{figure}[!ht]
\centering
\includegraphics[width=0.65\textwidth]{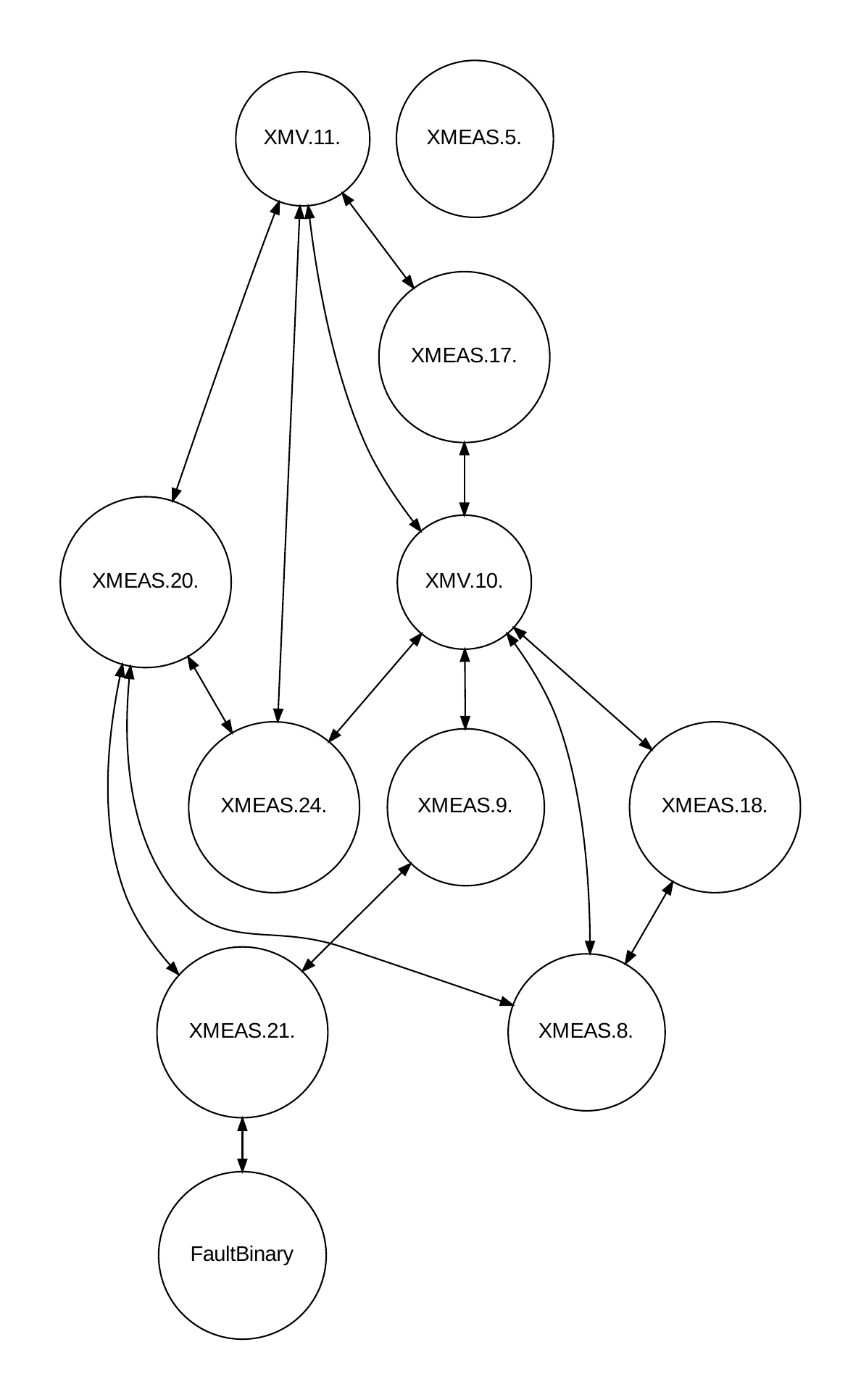}
\caption{DAG generated by the RFCI algorithm for the 10-variable subset. This representation shows XMEAS.21 (Reactor cooling water outlet temperature) as the direct precursor to fault conditions, with bidirectional relationships between cooling variables (XMV.10, XMV.11) indicating complex feedback mechanisms in the industrial process.}
\label{fig:dag-rfci}
\end{figure}

\begin{figure}[!ht]
\centering
\includegraphics[width=0.65\textwidth]{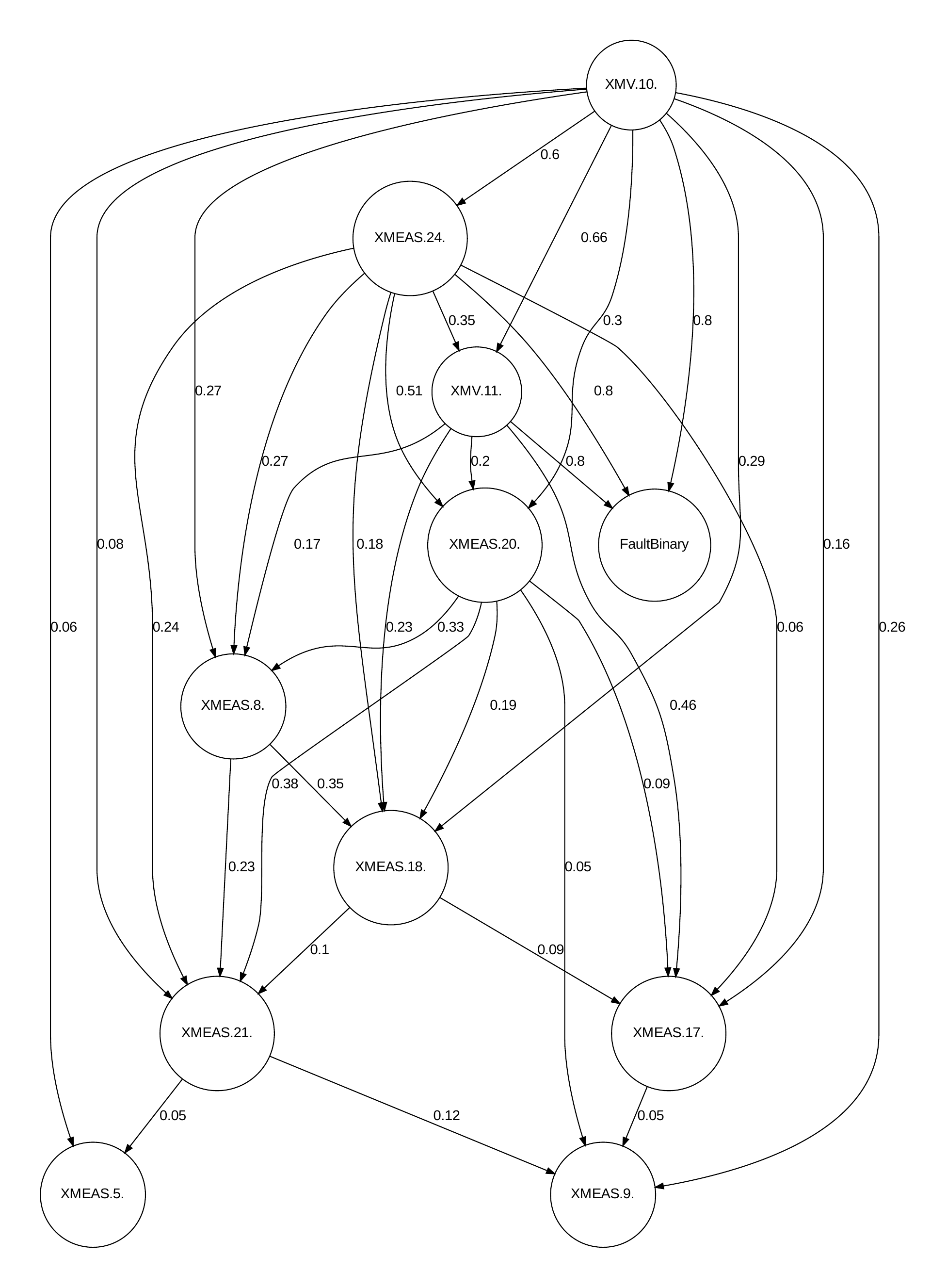}
\caption{DAG generated by the NOTEARS algorithm for the 10-variable subset. Edge weights indicate causal strength, with XMV.10 (Reactor cooling water flow) serving as the primary root node with direct, high-strength connections (weight 0.8) to fault conditions.}
\label{fig:dag-notears}
\end{figure}

\begin{figure}[!ht]
\centering
\includegraphics[width=0.65\textwidth]{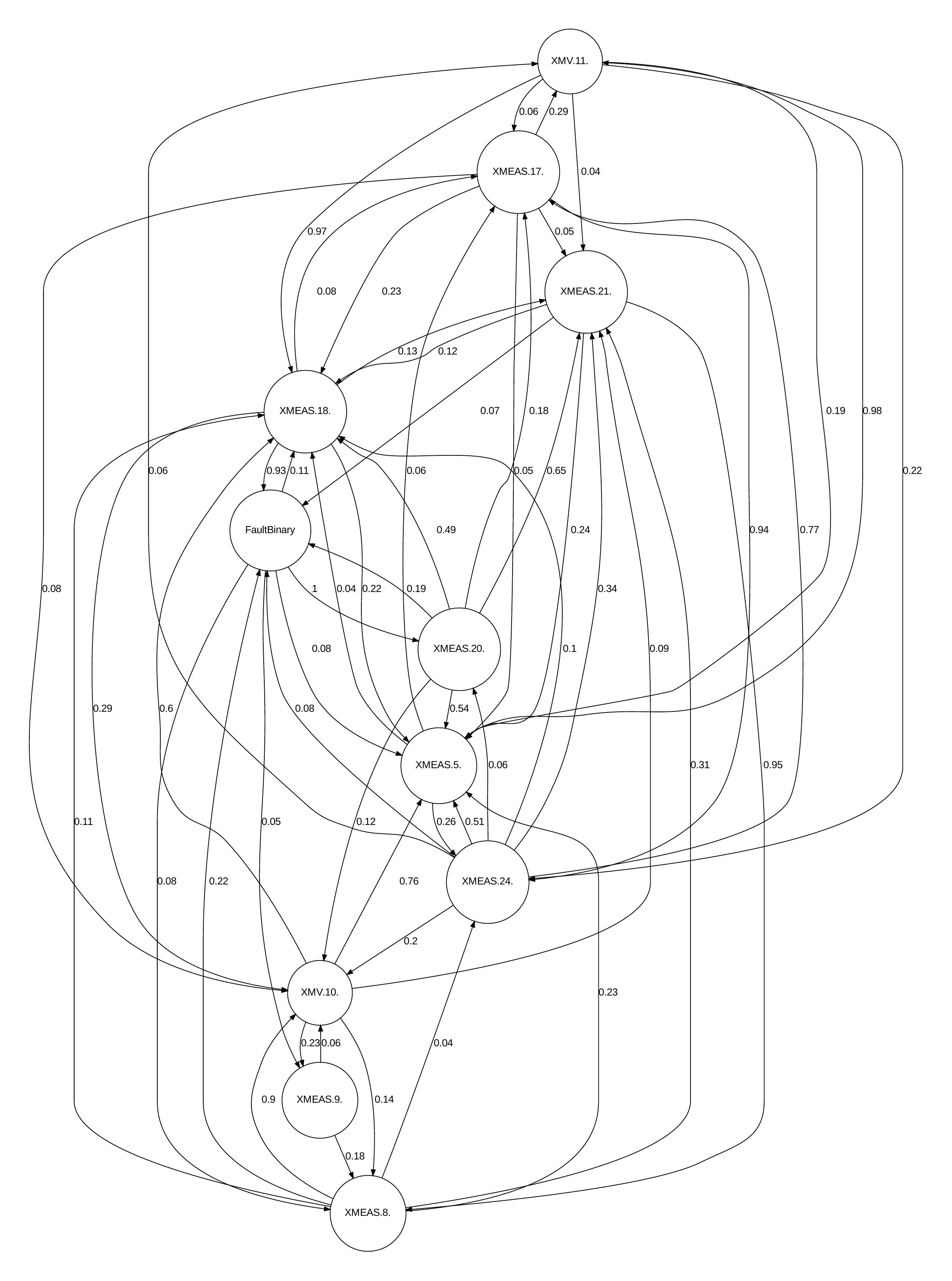}
\caption{DAG generated by the LINGAM algorithm for the 10-variable subset. This densely connected network with weighted edges reveals XMEAS.18 (Stripper temperature) as having the strongest direct influence on fault occurrence (weight 0.93) and XMEAS.5 (Recycle flow) strongly influencing XMV.11 (weight 0.98).}
\label{fig:dag-lingam}
\end{figure}

Figs. \ref{fig:dag-pc} through \ref{fig:dag-lingam} present the DAG structures generated by each causal discovery algorithm. Despite the methodological differences between algorithms, several consistent patterns emerged that have significant implications for industrial process monitoring. The critical role of cooling systems stands evident across all algorithms, with XMV.10 (Reactor cooling water flow) and XMV.11 (Condenser cooling water flow) consistently appearing as central nodes with extensive causal connections. The significance of stripper operations is another consistent finding, with XMEAS.17 (Stripper underflow) and XMEAS.18 (Stripper temperature) emerging as causally significant across all algorithms.

\section{Discussion}
\label{sec:discussion}

Our integrated framework offers several practical implications for industrial fault detection and process control. Variables identified as both predictively important and causally influential should be prioritized in industrial monitoring systems. This dual significance enables a focused approach to sensor deployment and maintenance scheduling, reducing resource demands while targeting the most critical process components. The strong causal role of manipulated variables like XMV.10 suggests specific opportunities for enhanced control strategies, while our results demonstrate that fault detection systems can operate effectively with reduced input dimensionality.

The framework novelty lies in its combination of machine learning interpretability with causal reasoning, advancing automated methodologies for industrial engineering. However, the computational complexity of causal discovery limits real-time applicability, suggesting future work on scalable algorithms or parallel computing. Temporal causal analysis could further capture dynamic fault propagation, enhancing predictive capabilities.

\section{Conclusion}
\label{sec:conclusion}

This study presents a novel computational framework for industrial fault detection by integrating SHAP-based feature selection with causal analysis via Directed Acyclic Graphs (DAGs), applied to the Tennessee Eastman Process (TEP) dataset. Our approach addresses high dimensionality and imbalanced data challenges, achieving an 80\% reduction in input variables (from 52 to 10) while maintaining MLP model accuracy at 0.872 and improving balanced accuracy from 0.650 to 0.711. In addition, the most known algorithms to causal discovery, PC, FCI, RFCI, NOTEARS and LINGAM were employed to generate DAGs with reduced variable sets obtained by SHAP. By combining SHAP's predictive interpretability with DAGs' causal insights, we uncover key fault mechanisms, consistently identifying cooling variables and stripper parameters as critical nodes across five causal discovery algorithms. These findings enable targeted monitoring, reducing sensor and computational demands while providing actionable insights for process control.

The framework's novelty lies in its combination of machine learning interpretability with causal reasoning, advancing computerized methodologies for industrial engineering. However, the computational complexity of causal discovery limits real-time applicability, suggesting future work on scalable algorithms or parallel computing. Temporal causal analysis could further capture dynamic fault propagation, enhancing predictive capabilities. By bridging predictive accuracy with causal understanding, this framework contributes to smarter, more reliable industrial systems, paving the way for broader applications in complex manufacturing environments.

\section*{Acknowledgements}

The authors thank the Brazilian research agencies FAPES and CNPq. R. A. Krohling thanks CNPq for financial support under grant no. 302021/2025-6.

\bibliographystyle{apalike}
\bibliography{references}


%
%
%
%
%

\end{document}